# SD-SLAM: A Semantic SLAM Approach for Dynamic Scenes Based on LiDAR Point Clouds

Feiya Li[1], Chunyun Fu[1]*, Dongye Sun[1], Jian Li[2], Jianwen Wang[2]

*1. College of Mechanical and Vehicle Engineering, Chongqing University, Chongqing, China*
*2. State Key Laboratory of Intelligent Vehicle Safety Technology, Chongqing Changan Automobile Co., Ltd., Chongqing, China*
*Corresponding Author's Email: fuchunyun@cqu.edu.cn*

**Abstract:** Point cloud maps generated via LiDAR sensors using extensive remotely sensed data are commonly used by autonomous vehicles and robots for localization and navigation. However, dynamic objects contained in point cloud maps not only downgrade localization accuracy and navigation performance but also jeopardize the map quality. In response to this challenge, we propose in this paper a novel semantic SLAM approach for dynamic scenes based on LiDAR point clouds, referred to as SD-SLAM hereafter. The main contributions of this work are in three aspects: 1) introducing a semantic SLAM framework dedicatedly for dynamic scenes based on LiDAR point clouds, 2) Employing semantics and Kalman filtering to effectively differentiate between dynamic and semi-static landmarks, and 3) Making full use of semi-static and pure static landmarks with semantic information in the SD-SLAM process to improve localization and mapping performance. To evaluate the proposed SD-SLAM, tests were conducted using the widely adopted KITTI odometry dataset. Results demonstrate that the proposed SD-SLAM effectively mitigates the adverse effects of dynamic objects on SLAM, improving vehicle localization and mapping performance in dynamic scenes, and simultaneously constructing a static semantic map with multiple semantic classes for enhanced environment understanding.

## 1. Introduction

Simultaneous Localization and Mapping (SLAM) has been extensively investigated in the field of robotics and autonomous driving [1]. Despite the emergence of some effective SLAM solutions such as [2-4], most existing SLAM solutions still rely on the assumption of a static world. However, in real environments, moving objects are ubiquitous and dynamic scenes are almost inevitable, which leads to performance deterioration or even failure of the existing SLAM algorithms. Moreover, the existing mapping methods have been largely based on traditional map types, such as geometric feature maps [5-7] and occupancy grid maps [8-10]. Although these maps present respective advantages in different SLAM applications, they still suffer from some obvious shortcomings such as inability to adapt to environment changes, inadequate environment understanding (e.g., inability to identify and classify objects), and ineffective human-machine interaction.

Aiming to address the above limitations of the existing SLAM solutions, this study proposes a novel semantic LiDAR SLAM approach for dynamic scenes. Our method employs a Fully Convolutional Neural (FCN) Network [11] for semantic segmentation, and the resulting segmented outputs are further clustered by means of DBSCAN [12] to achieve instance segmentation. To accurately differentiate between dynamic and stationary objects, we introduce semantics and Kalman filtering into our strategy to distinguish the motion states of landmarks. By this means, three types of landmarks can be identified, including dynamic landmarks, semi-static landmarks (stationary at the moment of detection, but may exhibit motion over time), and pure static landmarks. On this basis, not only can the adverse influence of dynamic landmarks on pose estimation be eliminated, but also greater number of landmarks (both semi-static and pure static landmarks) become available for registration to enhance localization performance. Then, the LiDAR odometry proposed in our previous work – L-LO [13] – is again utilized in this study to provide pose estimation results, and its objective function is solved by means of an optimized particle swarm optimization (PSO) algorithm [14] for localization and mapping accuracy enhancement. In terms of loop closure, the BoW3D algorithm [15] is refined and improved in this study to rectify accumulated localization errors. The proposed LiDAR SLAM method effectively counters the overreliance on the static environment assumption, and significantly enhances localization performance in dynamic scenes. Besides, the proposed method simultaneously constructs a static map, in which landmarks are categorized into multiple semantic classes, thereby highlighting its comprehensive environmental understanding.

The primary contributions of this paper include:
- Introducing a novel semantic SLAM framework dedicatedly for dynamic scenes based on LiDAR point clouds;
- Integrating semantics and Kalman filtering to effectively differentiate between dynamic and semi-static landmarks in the scenes;
- Making full use of semi-static and pure static landmarks with semantic information in the SD-SLAM process to enhance localization and mapping performance.

In summary, the proposed SD-SLAM effectively suppresses the adverse impact of dynamic objects on SLAM performance, enhances the localization accuracy, and provides static semantic maps with multiple semantic classes.

## 2. Related works

In recent years, significant advancements have been observed in various areas related to localization and mapping based on 3D point clouds. To provide a necessary background for the proposed method, in this section we offer a concise review of pertinent literature, primarily emphasizing: 1) LiDAR-based localization and mapping, and 2) LiDAR SLAM in dynamic scenes.

### 2.1 LiDAR-based localization and mapping

Zhang et al. [2] proposed a well-known lidar odometry and mapping method, commonly referred to as LOAM, to address the real-time localization and mapping problem. This method has consistently held high rankings on the KITTI odometry leaderboard for several years, and it has become a benchmark in the field of LiDAR SLAM. This approach extracts point features from the original point cloud and classifies these features into two types based on point curvature: edge points and planar points. Using these two types of point features, the pose of the LiDAR is estimated by means of the Levenberg-Marquardt optimization algorithm. Wang et al. [16] proposed a variant of the original LOAM, referred to as Fast LiDAR Odometry and Mapping (F-LOAM). F-LOAM adopts a non-iterative two-stage distortion compensation approach to reduce computational cost and improve computational efficiency. Shan et al. [17] proposed another variant of the original LOAM, which is a tightly-coupled lidar inertial odometry via smoothing and mapping (LIO-SAM). The primary enhancement of this algorithm over LOAM is its superior capability in loop closure detection and the integration of other absolute measurements, such as GPS, for pose refinement. Oelsch et al. [18] introduced an improved odometry algorithm based on LOAM, known as R-LOAM. In addition to traditional point features, mesh features are also extracted from a reference object in this algorithm. By integrating the correspondence between points and meshes into a unified optimization framework, the localization accuracy of R-LOAM is improved. Yokozuka et al. [19] developed an improved version of iterative closest point (ICP) approach based on symmetric Kullback-Leibler divergence (KL-Divergence) for point cloud registration. This method allocates 3D points measured by LiDAR to voxel grids, and depending on how 3D points are scattered in voxel grids, the points in each voxel are represented by an individual Gaussian distribution. This approximation significantly reduces the number of points used for registration, thereby shortening computation time while maintaining high matching accuracy. Based on Normal Distributions Transform (NDT) and LOAM, Chen et al. [4] put forward a LiDAR odometry and mapping method named NDT-LOAM. The improvements of NDT-LOAM over NDT and LOAM include incorporating a weighted NDT method, and performing Local Feature Adjustment (LFA) through feature correspondence and pose adjustment to increase the accuracy of point cloud registration and pose estimation. Park and Yi [6] proposed a least squares matching algorithm based on line segment features for robotic SLAM applications. In this approach, maps are represented in the form of line segment features, and feature registration is completed through a line segment feature matching algorithm. By this means, not only is the memory space required for map storage reduced, but also the feature registration efficiency is enhanced. To address the measurement errors present in raw LiDAR data, Li and Amanda [20] proposed a SLAM approach based on planar features and point cloud intensity information. This method not only utilizes geometric features extracted from point clouds, but also incorporates point intensity information for pose estimation. Besides, back-end optimization is employed as well in this approach to help improve the accuracy of localization and mapping.

As we have seen, the aforementioned algorithms predominantly rely on the static environment assumption. However, real-world scenarios often encompass dynamic objects such as pedestrians and moving vehicles. These mobile objects can result in erroneous data associations during feature registration, detrimentally impacting the performance of pose estimation and mapping. To address this problem, solutions have been proposed in the literature to deal with the adverse effects of moving objects in dynamic scenes.

### 2.2 LiDAR SLAM in dynamic scenes

Landmarks can be broadly categorized based on their motion states, when being detected by sensors, into two primary classes: 1) Dynamic landmarks, exemplified by moving vehicles and bicycles; and 2) Static landmarks, exemplified by structures, traffic lights, temporarily parked vehicles and bicycles. The static landmarks can be further divided into two sub-categories based on their semantic information: a) Pure static landmarks, such as buildings and traffic lights; and b) Semi-static landmarks, which are stationary at the moment of detection, but may exhibit motion over time, such as temporarily parked vehicles and bicycles.

To deal with the adverse effects of the moving objects on localization and mapping performance, various solutions have been proposed in the literature. Following the above categorization of landmarks, the existing SLAM methods which take into account object motions can be also categorized into two types: 1) SLAM methods based on pure static landmarks, and 2) SLAM methods based on static landmarks (pure static and semi-static landmarks).

In this type of methods, only pure static landmarks are employed in the SLAM process, and the semi-static landmarks which are temporarily stationary at the time of detection are excluded. Chen et al. [21] proposed a semantic LiDAR SLAM approach named SuMa++. This approach employs a FCN network for semantic segmentation of point clouds, which is subsequently utilized to filter out dynamic and semi-static objects. On this basis, pure static landmarks are employed for registration based on ICP matching to achieve pose estimation, along with the generation of a high-quality semantic map. Jian et al. [22] utilized the semantic information generated by SegNet [23] to remove dynamic landmarks and semi-static landmarks from RGB images. After landmark removal, the remaining semantic information in RGB images is fused with LiDAR measurements, and then pose estimation is accomplished using the fused information. This method effectively mitigates the adverse impact of dynamic and semi-static landmarks on vehicle positioning in dynamic scenes. Moreno et al. [24] studied the effect of point cloud semantic segmentation on LiDAR odometry performance. In their investigation, the segmented point clouds are processed based on different semantic objects, such as removing dynamic landmarks, excluding vehicles, eliminating distant points, discarding ground points, and retaining pure static landmarks. By processing different semantic objects, their impact on LiDAR odometry is then revealed.

Apart from solutions based on pure static landmarks, some SLAM methods were designed based on more "relaxed" conditions. Namely, in these approaches not only pure static landmarks but also semi-static landmarks are employed. Pfreundschuh et al. [25] introduced a LiDAR SLAM algorithm tailored for real-time detection and management of moving objects. This method employs the 3D-MiniNet neural network to detect dynamic landmarks, and automatically labels these landmarks based on occupancy grids. Subsequently, both pure static and semi-static landmarks are utilized to localize the robot. Cong et al. [26] proposed a robust mobile robot mapping system that effectively integrates three types of algorithms, including LiDAR SLAM, dynamic object detection and removal, and probabilistic map fusion. On the one hand, the front-end of this system achieves real-time dynamic object detection and tracking based on point cloud density, which facilitates pose estimation of the LiDAR SLAM module. On the other hand, the back-end uses a probabilistic map fusion method to filter out unreliable observations resulting from motion blur. This integrated system ensures not only high-precision robot localization but also high-quality point cloud map construction in dynamic environments.

Li et al. [27] proposed to project point cloud data onto RGB images to generate dense depth images. By integrating these depth images with RGB sequences, depth residual is computed to exclude dynamic landmarks and retain static and semi-static landmarks. Using the resulting static environment, enhanced robot pose estimation accuracy can be then achieved. Li et al. [28] designed a LiDAR SLAM algorithm capable of dynamic object filtering. This approach employs the 3D-MiniNet neural network to detect and remove point features on dynamic objects. With dynamic features removed, static point features are then extracted from the filtered point cloud. By conducting static feature registration, the vehicle pose can be accurately estimated.

As discussed above, the current LiDAR SLAM approaches merely relying on pure static landmarks directly discard semi-static and dynamic landmarks in the environment. While this strategy is straightforward and easy for implementation, it must be noted that the removal of semi-static landmarks may reduce the number of useful landmarks, thereby adversely affecting localization and mapping performance. Additionally, the existing LiDAR SLAM methods utilizing both pure static and semi-static landmarks do not incorporate semantic information during registration, which poses inevitable risk of incorrect feature matches. The omission of semantics also results in maps that lack classification information.

In the following contents, we shall demonstrate the necessity and significance of devising an effective approach to differentiate dynamic, semi-static and pure static landmarks, and utilizing semi-static and pure static landmarks with semantic information in SLAM applications. Using our proposed approach, adverse effects caused by dynamic objects can be suppressed, and the overall localization and mapping performance in dynamic scenes can be effectively enhanced.

## 3. Methodology

This paper proposes a novel semantic LiDAR SLAM framework for dynamic scenes, as illustrated in Fig. 1. The overall framework consists of six components: 1) Instance Segmentation: This module segments the point cloud raw data, and endows the point cloud with semantic and instance attributes. 2) Preliminary Pose Estimation: This component employs the segmented landmarks in consecutive frames for registration, thereby producing an initial estimate of vehicle pose. 3) Landmark Motion State Identification: Using the preliminary pose estimation results, the motion states (i.e. position and velocity) of landmarks can be effectively predicted and tracked based on a Kalman filtering framework. By further leveraging the semantics and motion states of landmarks, dynamic and semi-static landmarks can be identified. 4) Precise Pose Estimation: In this module, a more sophisticated registration approach is proposed to achieve more precise six degree-of-freedom (DOF) pose estimation, based on the results of preliminary pose estimation and landmark motion state identification. 5) Loop Closure: To ensure global consistency of the localization and mapping results, an enhanced BoW3D algorithm [15] is employed to correct trajectory drift and map duplication resulting from error accumulation. 6) Pure Static Semantic Mapping: In this module, pure static landmarks are employed to construct an accurate, detailed 3D semantic map with essential pure static attributes for navigation and path planning. The details of each component in our proposed framework are explained in the following sections.

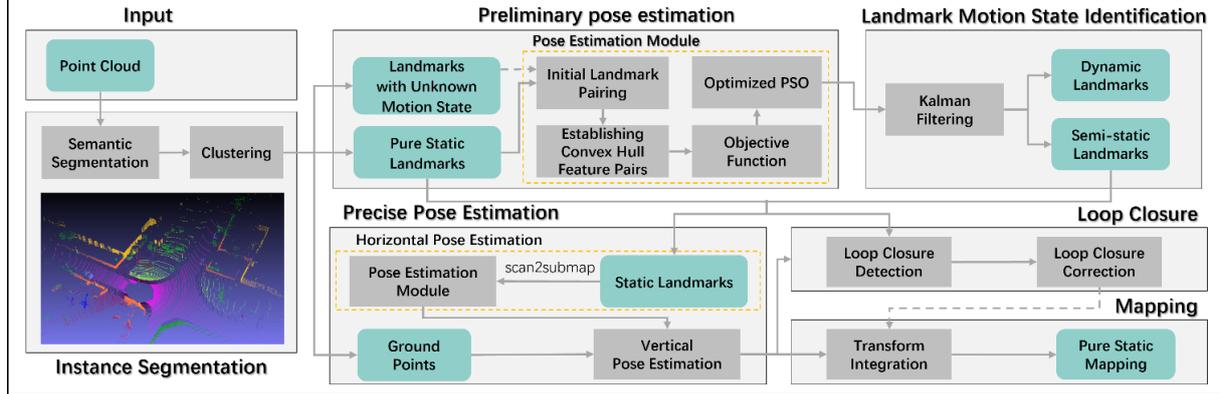

**Figure 1.** Proposed semantic LiDAR SLAM framework for dynamic scenes.

### 3.1 Point cloud instance segmentation

The purpose of point cloud instance segmentation is to assign a unique instance ID to each point in the point cloud to differentiate individual objects [29]. In this study, semantic segmentation is first performed to endow point cloud with semantic information by means of RangeNet++ [11]. Subsequently, 3D points with identical semantic labels are further clustered using the DBSCAN algorithm [12]. By this means, 3D points that belong to the same instance are partitioned from the rest and given the same ID, which accomplishes instance segmentation.

#### 3.1.1 Semantic segmentation

In this study, RangeNet++ [11] is employed to perform point cloud semantic segmentation. This network employs range images as an intermediate representation, integrating them with a convolutional neural network (CNN) tailored to a LiDAR sensor model, thereby achieving semantic segmentation of LiDAR point cloud.

This method improves computational efficiency by applying 2D convolution on range images. However, utilizing range images as an intermediate representation also introduces new issues such as discretization errors and blurry CNN outputs. To address these issues, this method introduces a GPU-based post-processing algorithm to enhance the accuracy of semantic segmentation.

Although RangeNet++ is adept at semantic segmentation of 3D point clouds, it does not facilitate instance segmentation. Therefore, in the following section, we introduce how point cloud instances can be obtained by clustering 3D points with the same semantics.

#### 3.1.2 Semantic segmentation

After conducting sematic segmentation, the point cloud with semantic information is then clustered using the DBSCAN algorithm [12, 30], thereby achieving point cloud instance segmentation. The DBSCAN algorithm achieves clustering based on the density distribution of points. Specifically, the operation of this algorithm hinges on two key parameters: local neighborhood radius, which defines the region centreed at a point, and minimum points in the neighborhood, which determines whether a point can be designated as a core point.

An apparent advantage of the DBSCAN algorithm is its ability of removing noise points while performing point clustering. It is also worth noting that, for different types of point clusters, local neighborhood radius and minimum points in the neighborhood can be both made adaptive to achieve more effective clustering and noise removal.

Based on the semantic information obtained from RangeNet++, in this study, invalid points such as unlabeled points (without semantics) and outlier points are first removed. Then, points that correspond to structures on the ground (such as roads, sidewalks and lane markings) can be easily identified. After the above steps, two types of point clusters are remained, i.e. clusters of pure static landmarks and clusters of landmarks with unknown motion states. These two types of point clusters are illustrated in Figs. 2(a) and 2(b), respectively.

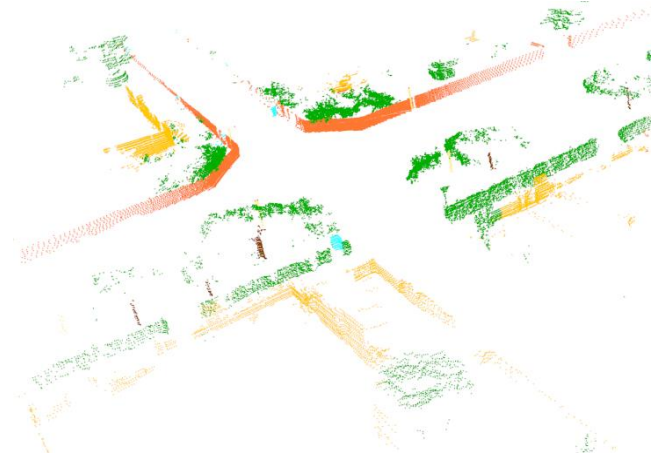

(a) Point clusters of pure static landmarks.

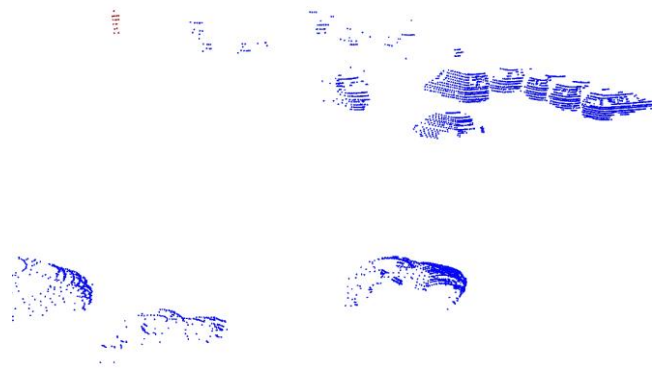

(b) Point clusters of landmarks with unknown motion states.

**Figure 2.** An illustration of point clusters of pure static landmarks and landmarks with unknown motion states. In sub-figure (a), pale yellow denotes buildings, dark green indicates vegetation, deep brown signifies trunks, bright yellow represents poles, red highlights traffic signs, and cyan corresponds to other objects. In sub-figure (b), blue symbolizes vehicles, while deep red designates persons.

When using the DBSCAN algorithm for clustering pure static landmarks, suitable key parameters are carefully selected according to the semantic information of different objects. Specifically, larger values of local neighborhood radii and minimum neighborhood points are used for larger objects such as buildings, sidewalks, and fences; smaller thresholds are employed for smaller objects like trunks, poles, and traffic signs. Similarly, when clustering landmarks with unknown motion states, appropriate key parameters are also selected based on different objects' semantics. Specifically, smaller thresholds are used for smaller objects like persons, bicyclists, and motorcyclists; medium-level thresholds are assigned to medium-sized objects such as cars, bicycles, and motorcycles; and larger thresholds are utilized for larger objects like buses, on-rail vehicles, and trucks.

During the course of vehicle pose estimation, relying solely on pure static landmarks may lead to inaccurate estimation results as the static landmarks can be too few to provide sufficient information for landmark registration. Hence, an intuitive remedy for this shortcoming is to involve more landmarks in the localization process. However, simply utilizing both pure static landmarks and landmarks with unknown motion states could compromise the accuracy of pose estimation, if the latter landmarks include dynamic objects.

Taking above into consideration, this study proposes an approach to determine the status of landmarks with unknown motion states by means of semantics and Kalman filtering. With this proposed approach onboard, pure static and semi-static landmarks can be both employed in the localization process to provide more precise pose estimation results. It must be pointed out that before identifying the motion status of landmarks, it is essential to first find out the motion status of the ego vehicle. Hence, in this study a preliminary pose estimation step is conducted for the ego vehicle, utilizing both pure static landmarks and landmarks with unknown motion states.

### 3.2 Preliminary pose estimation

In the preliminary pose estimation stage, we solely estimate the motion variations of the ego vehicle in the horizontal plane, i.e. translations in the $x$ and $y$ directions and rotation about the $z$-axis. The other motion variations of the ego vehicle such as roll and pitch motions are temporarily neglected in this stage. Therefore, we might as well define the vehicle pose variation between two consecutive frames as $t = [\Delta x, \Delta y, \Delta \theta]^T \in \mathbb{R}^{3 \times 1}$, where $\Delta x$ and $\Delta y$ denote the amount of position variations in the $x$ and $y$ directions respectively, and $\Delta \theta$ denotes the amount of yaw angle variation.

As mentioned previously, deficiency of pure static landmarks and existence of dynamic landmarks can lead to serious deterioration of localization performance. To mitigate the adverse effects caused by these issues, in this study we employ two different types of landmark inputs for preliminary point cloud registration, depending on the number of existing landmarks in the environment: 1) employing only pure static landmarks for registration when their quantity is adequate, and 2) utilizing all landmark types for registration if pure static landmarks are scarce or outnumbered by landmarks with unknown motion states. By this means, both accuracy and robustness of the preliminary pose estimation process can be ensured. Once the type of landmark inputs is determined, the following preliminary pose estimation process is executed to produce an initial estimate of vehicle pose.

### 3.2.1 Initial landmark pairing

To establish the correspondence between landmarks in two consecutive frames (i.e., frame $k-1$ and frame $k$), in this study we introduce an initial landmark pairing approach. Firstly, the geometric centre of each landmark, also referred to as the landmark centre [13], is computed. As illustrated in Fig. 3, for each landmark centre in frame $k-1$ (blue dots), we evaluate its distance to all landmark centres in frame $k$ (red dots) and choose the closest one as its match in frame $k$. If the minimum distance between two landmark centres is less than the average of the minimum distances between all matched landmark centres, then these two landmarks are considered to satisfy the "distance criterion" of landmark centres.

To further enhance accuracy of landmark pairing, apart from satisfying the above mentioned "distance criterion" of landmark centres, the semantic information of the two landmarks to be paired must be consistent, which is employed as the second criterion (also referred to as the "semantic criterion") for landmark pairing. Employing both "distance criterion" and "semantic criterion" effectively reduce the risk of landmark mismatches, as demonstrated in Fig. 3. Landmarks from frame $k-1$ and frame $k$ that satisfy both criteria are then referred to as "landmark pairs".

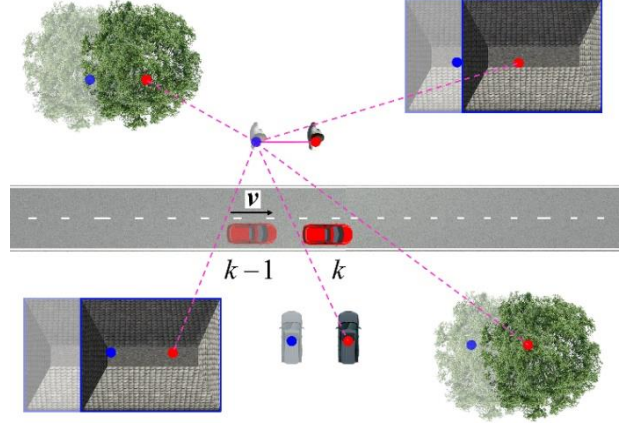

**Figure 3.** Landmark pairing using distance and semantic criteria in consecutive frames. Landmark centres in frame $k-1$ are represented by blue dots, while those in frame $k$ are depicted as red dots. Dashed pink lines indicate the evaluation of distances between landmark centres from both frames. Solid lines denote landmarks that satisfy the distance criterion and semantic criterion.

### 3.2.2 Establishing convex hulls for landmark pairs

In this study, convex hulls are used to represent landmarks' geometric shapes and sizes. To establish convex hulls of landmarks, we first perform vertical layering for each pair of landmarks. The threshold for vertical layering is determined by the average height of the landmark pair in the vertical direction. After layering is completed, the resulting 3D point clusters of the landmark pair are projected onto the horizontal plane to acquire the corresponding 2D point sets. For details of landmark layering and point projection, interested readers are referred to our previous work [13]. Then, the Graham's Scan algorithm [31] is applied to construct convex hulls of these 2D point sets. Apparently, each landmark pair is composed of two landmarks (one from frame $k-1$ and the other from frame $k$), each of which contains two layers – the upper layer and the lower layer. In other words, each landmark pair includes four individual convex hulls, as shown in Fig. 4.

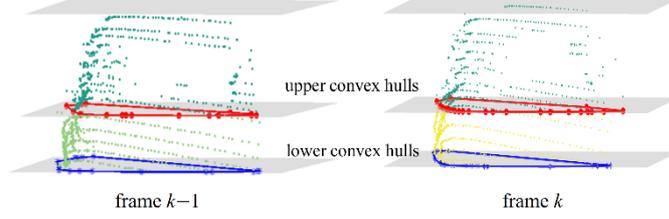

**Figure 4.** Upper and lower convex hulls for a landmark pair.

Once convex hulls are established, the similarity (in terms of shape and size) between two convex hulls in the same layer is evaluated by means of a comprehensive similarity index proposed in [13]. For each pair of landmarks, we have two pairs of convex hulls, one pair in the upper layer and the other pair in the lower layer (see Fig. 4). The above-mentioned comprehensive similarity index is employed to evaluate the similarity between the two convex hulls in each layer, in terms of both shape and dimension. The pair of convex hulls with higher similarity is termed the "convex hull features" of the corresponding landmark pair, which is then used as registration features to produce preliminary pose estimation results.

### 3.2.3 Objective function design

In this section, we introduce an objective function to quantify the level of overlap between the "convex hull features" established in the preceding section, for the purpose of pose estimation. By using this objective function, we are aiming to find out the optimal vehicle pose which maximizes the overlapping area between two convex hull features [13].

For two consecutive frames $k-1$ and $k$, we might as well denote the $i$-th pair of convex hull features as $\hat{S}_{k-1}^i$ and $\hat{S}_k^i$, respectively. Transforming $\hat{S}_k^i$ from frame $k$ to frame $k-1$, the overlapping area between $\hat{S}_{k-1}^i$ and $\hat{S}_k^i$ can be calculated as follows:

$$g_i(t) = \text{Area}((S_k^i \otimes t) \cap S_{k-1}^i) \qquad (1)$$

where Area(·) denotes the operation of computing polygon area, $\otimes$ acts as a coordinate transformation operator, and $g_i(t)$ denotes the overlapping area of the $i$-th pair of convex hull features after coordinate transformation. Note that this operation, $S_k^i \otimes t$, transforms convex hull $S_k^i$ from the vehicle coordinate system in frame $k$ to that in frame $k-1$, based on the vehicle pose change information $t$.

Based on the above formulation, the following objective function $G(t)$ for preliminary pose estimation can be devised:

$$G(t) = \underset{t}{\arg\max}\left(\sum_{i=1}^{N} g_i(t)\right) \qquad (2)$$

where $N$ represents the total number of matched pairs of convex hull features in frames $k-1$ and $k$. This objective function is formulated to provide a preliminary estimate of pose change $t$ by finding the maximum of total overlapping areas across $N$ pairs of convex hull features.

### 3.2.4 Solving objective function utilizing optimized PSO

We solve the above designed objective function using an optimized PSO algorithm [14]. In this algorithm, each particle has two attributes: position and velocity, where the position represents the "pose variation" and the velocity represents the "adjustment of pose variation". In every iteration, both position and velocity of each particle are updated. Firstly, the velocity of each particle is updated according to its personal best position as well as the global best position of the swarm. Then, the position of each particle is updated accordingly, based on the newly updated velocity information. By this means, the optimized PSO algorithm accomplishes efficient search of the feasible region and then finds the global optimal solution. Note that the position update is conducted after the velocity update, as the new velocity directs the particle to a new position in the search space. For more information on how this optimized PSO algorithm works, interested readers are referred to [14] and we do not repeat the details of this algorithm here due to page limits.

Utilizing the optimized PSO algorithm, a synergistic search considering both the global optimum and the personal optimum is achieved, which enables effective preliminary pose estimation within the feasible region. However, the resulting preliminary pose estimate may still suffer performance deterioration due to the adverse effects of dynamic landmarks as well as limited number of pure state landmarks. To enhance localization accuracy in such scenarios, in the following section, semantic information and filtering algorithms are introduced in our SLAM framework to realize accurate identification of landmark status. By this means, we are able to find out semi-static landmarks and then use them in the SLAM process, together with pure static landmarks, to enhance the localization performance.

### 3.3 Motion State Recognition of Landmarks with Unknown Motion States

Landmarks with unknown motion states are composed of semi-static landmarks and dynamic landmarks. The former ones refer to landmarks that appear stationary when observed but may undergo changes in position or velocity after a period of time, such as vehicles temporarily parked at roadsides. In contrast, the latter landmarks continually change their positions or directions during continuous observations, exemplified by moving vehicles and pedestrians.

If a landmark is semi-static, apparently, its positions in the world coordinate system in frames $k-1$ and $k$ should be very close (theoretically the same). In contrast, dynamic landmarks, due to their inherent mobility, will exhibit larger deviations in positions at different time instances in the world coordinate system. On this basis, Kalman filtering is employed in this study for target tracking, in order to determine the landmark motion states.

For landmarks with unknown motion states, a motion model is established to predict the motion states (including position and velocity) of landmark centres, and a measurement model is constructed to update their states. Then, the landmark motion states can be determined using a Kalman filter, based on the established motion model and measurement model.

#### 3.3.1 Motion model and measurement model

In our Kalman filter design, a two-dimensional constant turn-rate model is utilized for prediction. The state vector used in this model is defined as $\boldsymbol{x} = [x, \dot{x}, y, \dot{y}]^\mathrm{T}$, where $x$ and $y$ denote the coordinates of a landmark centre. The process noise is defined as $\boldsymbol{w} = [w_x, w_y]^\mathrm{T}$, with $w_x$ and $w_y$ being zero-mean Gaussian white noise components. The turning rate is defined as $\omega = \Delta\theta/\tau$, where $\Delta\theta$ denotes the amount of heading angle change and $\tau$ represents the LiDAR sampling period. Based on the landmark centre state in frame $k-1$, we can predict the landmark centre state in frame $k$ using the following state prediction equation (motion model):

$$\boldsymbol{x}_k = \boldsymbol{F}\boldsymbol{x}_{k-1} + \boldsymbol{\Gamma}\boldsymbol{w}_{k-1} \qquad (3)$$

where $\boldsymbol{x}_k$ denotes the predicted state in frame $k$, $\boldsymbol{x}_{k-1}$ represents the state in frame $k-1$, $\boldsymbol{w}_{k-1}$ denotes the process noise in frame $k-1$, $\boldsymbol{F}$ indicates the state transition matrix with the following form:

$$\boldsymbol{F} = \begin{bmatrix} 1 & \dfrac{\sin\omega\tau}{\omega} & 0 & -\dfrac{1-\cos\omega\tau}{\omega} \\ 0 & \cos\omega\tau & 0 & -\sin\omega\tau \\ 0 & \dfrac{1-\cos\omega\tau}{\omega} & 1 & \dfrac{\sin\omega\tau}{\omega} \\ 0 & \sin\omega\tau & 0 & \cos\omega\tau \end{bmatrix}$$

and $\boldsymbol{\Gamma}$ stands for the noise matrix with the following form:

$$\boldsymbol{\Gamma} = \begin{bmatrix} \tau^2/2 & 0 \\ \tau & 0 \\ 0 & \tau^2/2 \\ 0 & \tau \end{bmatrix}$$

To rectify state prediction errors, a measurement model is constructed to map measurements into the state space and update the predicted states. In the update step, the position of landmark centre in the LiDAR coordinate system of frame $k$ is transformed into the world coordinate system using the preliminary pose estimation results, and this transformed position serves as the input to the measurement model. Accordingly, the measurement equation is defined as follows:

$$\boldsymbol{z}_k = \boldsymbol{H}\boldsymbol{x}_k + \boldsymbol{v}_k \qquad (4)$$

where $\boldsymbol{z}_k = [z_{x,k}, z_{y,k}]^\mathrm{T}$ represents the measurement in frame $k$, $\boldsymbol{v}_k = [v_{x,k}, v_{y,k}]^\mathrm{T}$ denotes the measurement noise (zero-mean Gaussian white noise) in frame $k$, and $\boldsymbol{H}$ signifies the measurement matrix, given by: $\boldsymbol{H} = \begin{bmatrix} 1 & 0 & 0 & 0 \\ 0 & 0 & 1 & 0 \end{bmatrix}$.

#### 3.3.2 Kalman filtering

In general, there exist multiple landmarks with unknown motion states within the environment. Accordingly, for each landmark, an individual Kalman filter is implemented to provide independent estimations of landmark centre states. By using the motion model and measurement model constructed in Section 3.3.1, these individual Kalman filters accomplish prediction and update of landmark centre states, which lays the foundation for the following landmark recognition task. Note that since we implement standard Kalman filters, for brevity purpose we do not repeat their detailed design process in this section.

### 3.3.3 Recognition of dynamic and semi-static landmarks

To differentiate between dynamic and semi-static landmarks, we employ the following recognition procedure: Firstly, for landmark $L_k$ in frame $k$, we find its globally nearest neighbor in frame $k-1$, $L_{k-1}$, with the same semantics. If $L_{k-1}$'s globally nearest neighbor is also $L_k$, then we investigate their centre state discrepancies, including position and velocity differences, between landmarks $L_k$ and $L_{k-1}$. If the centre state discrepancies between $L_k$ and $L_{k-1}$ are sufficiently small (i.e. less than respective thresholds), then landmark $L_k$ can be recognized as a semi-static landmark. On the contrary, if these discrepancies are large, then landmark $L_k$ should be identified as a dynamic landmark.

It should be noted that in the above procedure, different threshold values are set for landmarks with different semantics. This is intuitive because landmarks with different semantics normally have different motion patterns, which in turn requires suitable thresholds to suit their individual dynamics. In summary, the process of recognizing dynamic and semi-static landmarks can be formulated as follows:

$$D = \begin{cases} 1, & \text{same semantics and } (\|\boldsymbol{v}_{\text{diff}}\| < \|\boldsymbol{v}_{\max}\| \text{ and } \|\boldsymbol{d}_{\text{diff}}\| < \|\boldsymbol{d}_{\max}\|) \\ 0, & \text{same semantics and } (\|\boldsymbol{v}_{\text{diff}}\| \geq \|\boldsymbol{v}_{\max}\| \text{ or } \|\boldsymbol{d}_{\text{diff}}\| \geq \|\boldsymbol{d}_{\max}\|) \end{cases} \quad (5)$$

where $D$ denotes the attribute of the landmark to be recognized, with "1" indicating a semi-static landmark and "0" representing a dynamic landmark, $\|\boldsymbol{v}_{\text{diff}}\|$ denotes the velocity difference between $L_k$ and $L_{k-1}$, $\|\boldsymbol{v}_{\max}\|$ represents the upper bound (threshold) for $\|\boldsymbol{v}_{\text{diff}}\|$, $\|\boldsymbol{d}_{\text{diff}}\|$ signifies the position difference between $L_k$ and $L_{k-1}$, $\|\boldsymbol{d}_{\max}\|$ is the upper bound (threshold) for $\|\boldsymbol{d}_{\text{diff}}\|$, and the term "same semantics" indicates that the semantic labels of $L_k$ and $L_{k-1}$ must be identical.

To sum up, in the above proposed landmark recognition procedure, both semantics and state discrepancies are concurrently taken into account for differentiating between dynamic and semi-static landmarks. In the following precise pose estimation phase, estimation accuracy is enhanced by eliminating dynamic landmarks and incorporating semi-static ones for registration.

## 3.4 Precise pose estimation

The task of precise pose estimation can be divided into two parts: horizontal pose estimation and vertical pose estimation. As mentioned above, horizontal pose estimation involves estimation of translations along the *x*- and *y*-axes, as well as rotation around the *z*-axis, with respect to the LiDAR coordinate system. Vertical pose estimation includes estimation of translation along the *z*-axis as well as rotation around the *y*-axis, with respect to the LiDAR coordinate system. Note that in this study, within the six DOF vehicle motion, a null roll angle (zero rotation around the *x*-axis) is hypothesized to simplify our computation. This hypothesis is appropriate as roll angle plays an important role in vehicle handling dynamics rather than in SLAM applications.

### 3.4.1 Horizontal pose estimation

In the horizontal pose estimation phase, we employ an approach similar to that used in Section 3.2, with two important differences: 1) Only pure static and semi-static landmarks are used, and dynamic landmarks are excluded in the estimation process. Recall that in the preceding preliminary pose estimation phase, either pure static landmarks or all landmarks are utilized for pose estimation. In the precise estimation stage, the adverse impacts of scarcity of pure static landmarks and presence of dynamic landmarks are effectively suppressed. 2) The 'scan to local sub-map' registration method is used to calculate the horizontal pose, while in the preliminary pose estimation phase the 'scan to scan' approach is adopted. This registration method achieves initial pairing by matching landmarks in the current frame with those in the local map, and then convex hulls are extracted from the initially paired landmarks. Subsequently, an optimized PSO algorithm [14] is deployed to solve the optimal pose transformation. The above two measures taken in the precise pose estimation process enhances the accuracy of horizontal pose estimation.

### 3.4.2 Vertical pose estimation

This section elucidates the methods we used to estimate the vehicle's pose in the vertical plane, i.e. translation along the *z*-axis and rotation around the *y*-axis (pitch angle). The pitch angle is estimated by leveraging point clouds reflected from the ground surface in front of and behind the vehicle. By fitting planes to these point clouds and deriving normal vectors of the resulting fitted planes, the pitch angle of the vehicle with respect to its initial pose can be obtained. With the estimated pitch angle, yaw angle and horizontal positions, the vehicle's translation along the *z*-axis can subsequently be determined. For more details about vertical pose estimation, interested readers are referred to our previously paper [13].

### 3.5 Loop closure

Loop closure detection plays a pivotal role in ensuring global consistency in SLAM applications. Due to sensor noise and odometry errors, trajectory drift becomes increasingly serious as time elapses. To mitigate this issue, in this study we employ the improved bag-of-words in 3D (BoW3D) algorithm [15] for loop closure detection.

The original BoW3D algorithm proposed in [15] extracts directionally invariant 3D features, referred to as LinK3D features, to construct a bag-of-words model. This model selects three nearest key points around each feature point, and computes distances between corresponding descriptors to filter out moving objects in the surrounding environment. However, this strategy might possibly face challenges in certain situations, for instance, the LinK3D features extracted from semi-static and dynamic landmarks could potentially diminish the robustness of loop closure detection, due to the effects of dynamic properties of these landmarks.

To tackle the above limitation, in this paper, we propose an improved BoW3D algorithm to eliminate the potential adverse effects of dynamic and semi-static landmarks on loop closure detection. In our approach, LinK3D features are solely extracted from pure static landmarks, namely, dynamic and semi-static landmarks are not involved in this process. Besides, only features with identical semantic information are utilized for registration and solving loop closure constraints. By this means, not only can the adverse effects resulting from dynamic and semi-static landmarks be suppressed, but also the accuracy of landmark registration can be enhanced. As a result, we achieve more effective loop closure correction and map database update.

### 3.6 Mapping

The primary goal of mapping is to depict the environment in an appropriate manner and construct a global map that can be used for path planning and navigation. In this study, the vehicle's pose is determined either through transform integration or, when loop closure is detected, by optimizing the pose after loop closure. Utilizing the obtained pose, the ground points and pure static points (including pure static landmark and ground points) are transformed from the vehicle coordinate system into the world coordinate system, thereby leading to a global map in the form of raw point cloud data. Subsequently, this global map is converted to a voxel grid representation for lightweight purposes, with the semantics of landmarks remaining invariant and intact. In this final semantic global map, dynamic landmarks and semi-static landmarks are fully filtered out, and only pure static points are retained. By this means, obstacle-free areas are generated for subsequent path planning and navigation tasks.

## 4 Evaluation results

The effectiveness of our proposed method has been verified through a comprehensive performance evaluation, encompassing three distinct dimensions: i.e. vehicle localization, dynamic landmark detection, and static mapping (using pure static landmarks and ground points). This multifaceted assessment was conducted based on the widely recognized KITTI dataset [32]. All experiments were executed on a machine equipped with a 16 GB RAM (8GB×2), a 512 GB hard drive, and an NVIDIA GeForce MX150 graphics card. The operating system used for evaluation was Ubuntu 18.04.

### 4.1 Tests with KITTI Datasets

Evaluation was conducted utilizing sequences 00 and 02-10 from the KITTI Odometry dataset, a well-known benchmark collection of various autonomous driving scenarios, in comparison with over 10 typical odometry solutions in the existing literature. The point cloud data in this dataset was captured at a frequency of 10 Hz by a Velodyne HDL-64E S2 LiDAR, which encompasses various environments including urban roads, highways, and country roads, providing a rigorous test environment for assessing the effectiveness of our SLAM approach across realistic, dynamic scenarios.

#### 4.1.1 Evaluation on Vehicle Localization Performance

We evaluated the localization accuracy of our proposed SD-SLAM method on the KITTI dataset under two conditions: without loop closure and with loop closure.

For the former condition, the SD-SLAM's performance is benchmarked against an array of advanced algorithms, including MULLS-LO (*s1*) [33], MULLS-LO (*mc*) [33], ICP-Point2Point [34], CLS [35], A-LOAM w/o mapping, G-ICP [36], Velas et al. [37], LOAM-Velodyne [2], LoDoNet [38], LeGO-LOAM [39] and L-LO [13]. In the latter scenario, the localization accuracy of our SD-SLAM algorithm is compared with those of two recent solutions – S4-SLAM [40] and LiTAMIN2 [19].

The evaluation relies on two key performance metrics: the average translation error ($t_{rel}$) and the average rotation error ($r_{rel}$) [33], which are used as standard indicators by the KITTI dataset for quantifying the level of precision of SLAM methods. Table 1 presents a localization accuracy comparison between our proposed SD-SLAM and more than ten advanced SLAM methods in the literature, with or without loop closure condition.

Table 1 Localization accuracy comparison between SD-SLAM and existing methods with or without loop closure condition.

| Seq.<br>Method | 00 U* | 01 H | 02 C* | 03 C | 04 C | 05 C* | 06 U* | 07 U* | 08 U* | 09 C* | 10 C | Average |
|---|---|---|---|---|---|---|---|---|---|---|---|---|
| | $t_{rel}/r_{rel}$ | $t_{rel}/r_{rel}$ | $t_{rel}/r_{rel}$ | $t_{rel}/r_{rel}$ | $t_{rel}/r_{rel}$ | $t_{rel}/r_{rel}$ | $t_{rel}/r_{rel}$ | $t_{rel}/r_{rel}$ | $t_{rel}/r_{rel}$ | $t_{rel}/r_{rel}$ | $t_{rel}/r_{rel}$ | $t_{rel}/r_{rel}$ |
| MULLS-LO(*s1*) | 2.36/1.06 | 2.76/0.89 | 2.81/0.95 | 1.26/0.67 | 5.72/1.15 | 2.19/1.01 | 1.12/0.51 | 1.65/1.27 | 2.73/1.19 | 2.14/0.96 | 3.61/1.65 | 2.57/1.03 |
| ICP-Point2Point | 6.88/2.99 | 11.21/2.58 | 8.21/3.39 | 11.07/5.05 | 6.64/4.02 | 3.97/1.93 | 1.95/1.59 | 5.17/3.35 | 10.04/4.93 | 6.93/2.89 | 8.91/4.74 | 7.37/3.38 |
| CLS | 2.11/0.95 | 4.22/1.05 | 2.29/0.86 | 1.63/1.09 | 1.59/0.71 | 1.98/0.92 | 0.92/0.46 | 1.04/0.73 | 2.14/1.05 | 1.95/0.92 | 3.46/1.28 | 2.12/0.91 |
| A-LOAM w/o mapping | 4.08/1.69 | 3.31/0.92 | 7.33/2.51 | 4.31/2.11 | 1.60/1.13 | 4.09/1.68 | 1.03/0.52 | 2.89/1.8 | 4.82/2.08 | 5.76/1.85 | 3.61/1.76 | 3.89/1.64 |
| G-ICP | 1.29/0.64 | 4.39/0.91 | 2.53/0.77 | 1.68/1.08 | 3.76/1.07 | 1.02/0.54 | 0.92/0.46 | 0.64/0.45 | 1.58/0.75 | 1.97/0.77 | 1.31/0.62 | 1.92/0.73 |
| Velas et al. | 3.02/- | 4.44/- | 3.42/- | 4.94/- | 1.77/- | 2.35/- | 1.88/- | 1.77/- | 2.89/- | 4.94/- | 3.27/- | 3.15/- |
| LOAM Velodyne | 3.41/- | 6.54/- | 5.66/- | 1.64/- | 1.09/- | 1.32/- | 1.01/- | 1.26/- | 2.16/- | 1.44/- | 1.91/- | 2.49/- |
| LeGO-LOAM | 2.17/1.05 | 13.4/1.02 | 2.17/1.01 | 2.34/1.18 | 1.27/1.01 | 1.28/0.74 | 1.06/0.63 | 1.12/0.81 | 1.99/0.94 | 1.97/0.98 | 2.21/0.92 | 2.49/1.00 |
| LO-Net | 1.47/0.72 | 1.36/0.47 | 1.52/0.71 | 1.03/0.66 | 0.51/0.65 | 1.04/0.69 | 0.71/0.50 | 1.70/0.89 | 2.12/0.77 | 1.37/0.58 | 1.80/0.93 | 1.33/0.69 |
| HALFlow | 10.46/4.46 | 72.21/14.66 | 24.54/8.64 | 15.21/8.03 | 54.30/34.02 | 10.55/4.12 | 14.35/5.87 | 14.24/8.15 | 24.59/9.64 | 21.43/8.15 | 19.03/8.29 | 25.53/10.37 |
| HALFlow-Refine | 2.89/1.26 | 2.37/0.72 | 2.64/1.12 | 2.79/2.12 | 1.74/0.88 | 3.01/1.29 | 3.23/1.14 | 2.92/1.82 | 4.13/1.60 | 3.05/1.11 | 3.62/1.78 | 2.94/1.35 |
| L-LO | 0.91/0.81 | -/- | 2.35/0.87 | 2.92/0.34 | 2.28/0.42 | 0.82/0.31 | 0.66/0.37 | 0.69/0.26 | 1.37/0.58 | 1.96/0.94 | 1.90/0.75 | 1.59/0.57 |
| SD-SLAM | 0.71/0.77 | -/- | 2.16/0.73 | 1.94/0.31 | 1.84/0.26 | 0.52/0.21 | 0.58/0.28 | 0.53/0.23 | 0.86/0.40 | 1.85/0.89 | 1.61/0.46 | 1.26/0.44 |
| S4-SLAM * | 0.62/- | 1.11/- | 1.63/- | 0.82/- | 0.95/- | 0.50/- | 0.65/- | 0.60/- | 1.33/- | 1.05/- | 0.96/- | 0.93/- |
| LiTAMIN2 * | 0.70/0.28 | 2.10/0.46 | 0.98/0.32 | 0.96/0.48 | 1.05/0.52 | 0.45/0.25 | 0.59/0.34 | 0.44/0.32 | 0.95/0.29 | 0.69/0.40 | 0.80/0.47 | 0.88/0.38 |
| SD-SLAM * | 0.63/0.33 | -/- | 0.61/0.40 | 1.94/0.31 | 1.84/0.26 | 0.38/0.12 | 0.23/0.17 | 0.30/0.11 | 0.65/0.31 | 0.62/0.37 | 0.98/0.36 | 0.82/0.27 |

Note: in this table, "U" represents urban roads, "H" denotes highways, "C" indicates country roads, "*" signifies loop closure presence in the environment, and "-" indicates data unavailability. The last three methods with blue shaded background include a loop closure detection module, while all other methods are merely odometry solutions without loop closure detection. Data marked in red indicate the best performance of corresponding metrics. The performance data, $t_{rel}$ and $r_{rel}$, of the compared methods were acquired from published papers or publicly available sources (e.g. arXiv).

It is seen in this table that when only odometry was used (i.e. no loop closure detection was involved), our proposed SD-SLAM achieved the highest localization accuracy across most test sequences. Besides, we see in the last column that SD-SLAM also provided the lowest $t_{rel}$ and $r_{rel}$ on average, across all test sequences used in this study. The above results demonstrate the advantages of our proposed SD-SLAM over the competing methods, when loop closure detection was not included.

On the other hand, with loop closure detection on-board, SD-SLAM also provided the best localization performance for most test sequences, as shown by the data with blue shaded background. In addition, we see from the last column that SD-SLAM also achieved the lowest $t_{rel}$ and $r_{rel}$ on average. These comparative results validate the efficacy of SD-SLAM with loop closure detection on-board.

Overall, our proposed SD-SLAM has demonstrated superior vehicle localization performance in comparison with numerous advanced counterparts in the existing literature, regardless of the presence of loop closure detection.

#### 4.1.2 Evaluation on Dynamic Landmark Detection Performance

To validate the effectiveness of the proposed algorithm in terms of dynamic landmark detection, in this study we meticulously selected sequence 07 from the KITTI odometry dataset for verification. This sequence encompasses a variety of scenarios which includes abundant dynamic landmarks, providing a rich repository of both dynamic and static landmarks.

The landmark detection performance using data from sequence 07 is demonstrated in Fig. 5. Note that in this figure dynamic landmarks are denoted in red, while semi-static and pure static landmarks are marked in grey. We observe that our proposed algorithm consistently and stably detected dynamic landmarks in the environment, and inferred the locations of dynamic landmarks in the constructed map. In the meantime, spatial consistency of semi-static and pure static landmarks was successfully maintained. Effectiveness of the proposed method can be verified through comparison between the detected dynamic landmarks (seen from the bird's eye view) and the ground truth in the images, both of which are clearly shown in Fig. 5.

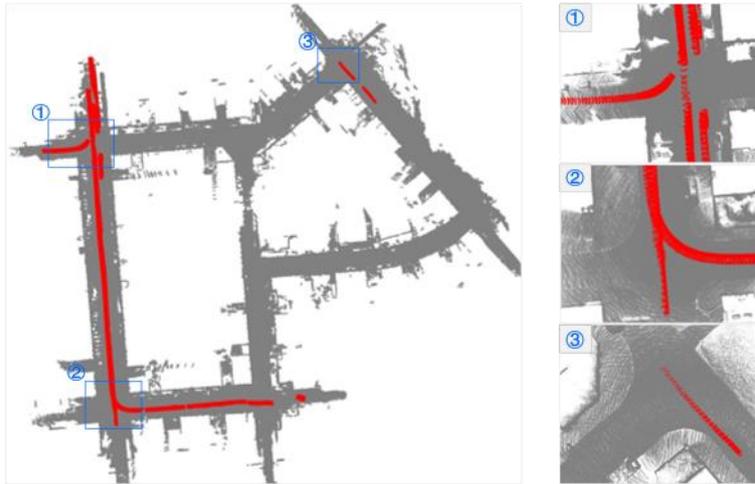

**Figure 5.** Performance of dynamic landmark detection as seen from the bird's eye view. Dynamic landmarks are denoted in red, while other landmarks are marked in grey. The three images on the right side show the ground truth of the three enlarged areas.

**4.1.3 Evaluation on Static Mapping Performance**

In this section, we demonstrate the performance of our proposed algorithm in terms of static map construction. Again, sequence 07 from the KITTI odometry dataset was used for this performance evaluation, due to its inclusion of abundant pure static landmarks and ground points.

In this study, only pure static points are selected for static map construction. This means that any point clusters semantically detected as dynamic or semi-static landmarks are excluded from the mapping process. The static mapping result is shown in Fig. 6, where different static instances are assigned a unique RGB color for better visualization. We observe that dynamic and semi-static landmarks are not included in the map, and only pure static landmarks (such as trunks, buildings, and traffic signs) are retained. By this means, the environmental information represented in this map is time-invariant, stable and consistent, thereby providing a solid foundation for applications relying on static environments such as SLAM, navigation and path planning.

**5. Conclusion and future works**

In this study, a novel semantic LiDAR SLAM framework is proposed, which is dedicated to addressing the challenges of dynamic environments, enhancing localization accuracy, and improving mapping performance. This framework is an effort in incorporating semantic information and Kalman filtering into SLAM, thereby effectively differentiating between dynamic and semi-static landmarks. Furthermore, semi-static and pure static landmarks with semantic information are made full use of in the SLAM process to enhance localization and mapping performance. In this approach, the same optimized PSO algorithm is employed at both the preliminary pose estimation stage and the precise pose estimation stage to solve the objective function.

Comparative evaluation on our SD-SLAM approach, conducted across a diverse array of scenarios, unequivocally demonstrates its superior localization performance when juxtaposed with existing methods. By addressing the traditional over-reliance on static environment assumptions, the proposed SD-SLAM manages to not only significantly boost localization accuracy in dynamic settings, but also effectively mitigate the adverse impacts of moving objects which has long plagued SLAM systems. Furthermore, this method goes beyond mere localization to concurrently construct a detailed static map, in which landmarks are categorized into multiple semantic classes. This feature underscores the method's ability to provide a comprehensive understanding of the surroundings. These attributes collectively demonstrate the clear advantages of the proposed SD-SLAM approach, showcasing its exceptional capability to navigate and map complex, ever-changing environments with abundant dynamic objects.

In our next step of investigations, focus will be placed on enhancements for localization robustness in highly degraded environments, by means of incorporating multi-modal sensor information.

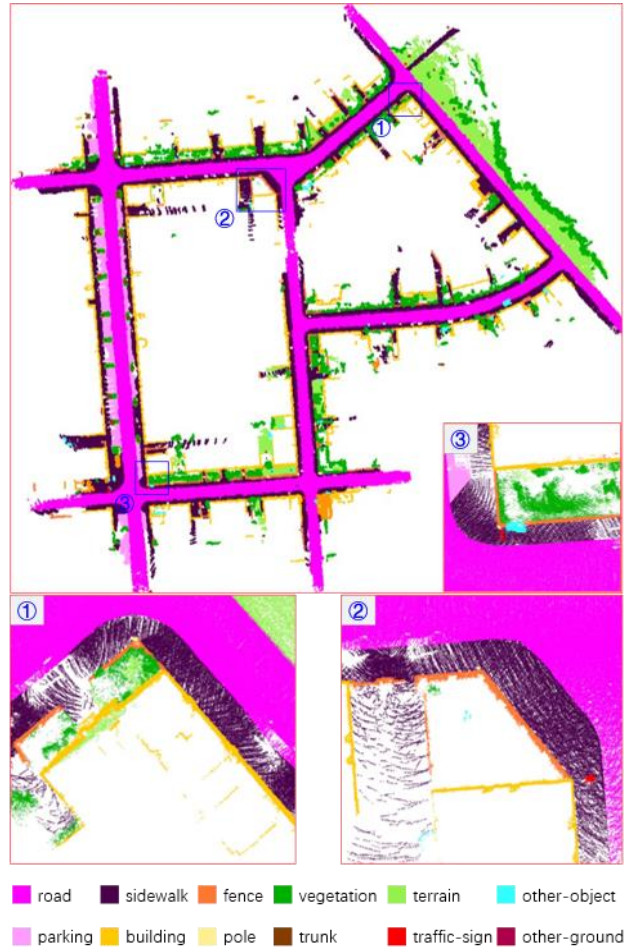

**Figure 6.** Mapping results with pure static landmarks and ground points using sequence 07 from the KITTI odometry dataset.

**Acknowledgements**